\pdfoutput=1

\documentclass[11pt]{article}

\usepackage[]{emnlp2021}

\usepackage{times}
\usepackage{latexsym}

\usepackage[T1]{fontenc}

\usepackage[utf8]{inputenc}

\usepackage{amssymb}
\usepackage{amsfonts}
\usepackage{amsmath}
\usepackage{enumitem}
\usepackage{booktabs}
\usepackage[ruled,vlined]{algorithm2e}

\usepackage{makecell}
\usepackage{multirow}
\usepackage{graphicx}
\usepackage{subcaption}
\usepackage{xcolor}
\definecolor{mygray}{gray}{0.92}
  {\list{}{\leftmargin=#1\rightmargin=#1}\item[]}%
  {\endlist}
\newcommand*{\affaddr}[1]{#1} 
\newcommand*{\affmark}[1][*]{\textsuperscript{#1}}
\newcommand*{\email}[1]{\texttt{#1}}

\usepackage{microtype}

\title{Aspect Sentiment Quad Prediction as Paraphrase Generation
\thanks{\hspace{1mm} Work done when Wenxuan Zhang was an intern at Alibaba. This work was supported by Alibaba Group through Alibaba Research Intern Program, and a grant from the Research Grant Council of the Hong Kong Special Administrative Region, China (Project Codes: 14204418).}
}

\author{
Wenxuan Zhang\affmark[1], Yang Deng\affmark[1], Xin Li\affmark[2], Yifei Yuan\affmark[1], Lidong Bing\affmark[2] and Wai Lam\affmark[1]\\
\affaddr{\affmark[1]The Chinese University of Hong Kong}\\
\affaddr{\affmark[2]DAMO Academy, Alibaba Group}\\
\email{\{wxzhang,ydeng,yfyuan,wlam\}@se.cuhk.edu.hk}\\
\email{\{xinting.lx,l.bing\}@alibaba-inc.com}\\
}

\begin{document}
\maketitle
\begin{abstract}
Aspect-based sentiment analysis (ABSA) has been extensively studied in recent years, which typically involves four fundamental sentiment elements, including the aspect category, aspect term, opinion term, and sentiment polarity. Existing studies usually consider the detection of partial sentiment elements, instead of predicting the four elements in one shot. In this work, we introduce the Aspect Sentiment Quad Prediction (ASQP) task, aiming to jointly detect all sentiment elements in quads for a given opinionated sentence, which can reveal a more comprehensive and complete aspect-level sentiment structure. We further propose a novel \textsc{Paraphrase} modeling paradigm to cast the ASQP task to a paraphrase generation process. On one hand, the generation formulation allows solving ASQP in an end-to-end manner, alleviating the potential error propagation in the pipeline solution. On the other hand, the semantics of the sentiment elements can be fully exploited by learning to generate them in the natural language form. Extensive experiments on benchmark datasets show the superiority of our proposed method and the capacity of cross-task transfer with the proposed unified \textsc{Paraphrase} modeling framework.
\end{abstract}

\section{Introduction}

As a fine-grained opinion mining problem, aspect-based sentiment analysis (ABSA) aims to analyse sentiment information at the aspect level \cite{liu-2012-absa, semeval14-absa}. Typically, four fundamental sentiment elements are involved in ABSA, including 1) \textit{aspect category} denoting the type of the concerned aspect; 2) \textit{aspect term} which can be either explicitly or implicitly mentioned in the given text; 3) \textit{opinion term} which describes the opinion towards the aspect; and 4) \textit{sentiment polarity} denoting the sentiment class. Given an example sentence ``\textit{The pasta is over-cooked!}'', the sentiment elements are ``\textit{food quality}'', ``\textit{pasta}'', ``\textit{over-cooked}'', and ``\textit{negative}'', respectively.

Due to its broad application scenarios, many research efforts have been made on ABSA to predict or extract those sentiment elements \cite{semeval14-absa, semeval15-absa, semeval16-absa}. Early studies focus on the prediction of a single element such as aspect term extraction \cite{emnlp15-ate, acl18-ate-xuhu}, aspect category detection \cite{aaai15-acp}, aspect sentiment classification based on either an aspect category \cite{emnlp16-ruder-asc-category, emnlp19-asc-category} or an aspect term \cite{emnlp18-asc}. 
More recent works propose to extract multiple associated sentiment elements at the same time \cite{acl21-gabsa}. For example, \citet{acl20-aope} consider the aspect and opinion term pairwise extraction; \citet{aaai20-robin} propose the aspect sentiment triplet extraction (ASTE) task to detect the (aspect term, opinion term, sentiment polarity) triplets; \citet{aaai20-tasd} handle the target aspect sentiment detection (TASD) task that jointly detects the aspect category, aspect term, and sentiment polarity. 

Despite their popularity, these ABSA tasks only attempt to perform partial prediction instead of providing a complete aspect-level sentiment picture, \textit{i.e.}, identifying the four sentiment elements in one shot. 
To this end, we introduce the aspect sentiment quad prediction (\textbf{ASQP}) task, aiming to predict all (aspect category, aspect term, opinion term, sentiment polarity) quads for a given opinionated sentence. This new task compensates for the drawbacks of previous tasks and helps us comprehensively understand user’s aspect-level opinions.

To tackle ASQP, one straightforward idea is to decouple the quad prediction problem into several sub-tasks and solve them in a pipeline manner. However, such multi-stage approaches would suffer severely from error propagation because the overall prediction performance hinges on the accuracy of every step \cite{aaai20-robin, acl20-aope}. 
Besides, the involved sub-tasks, which are usually formulated as either token-level or sequence-level classification problems, underutilize the rich semantic information of the label (\textit{i.e.}, the meaning of sentiment elements to be predicted) since they treat the labels as number indices during training. Intuitively, the aspect term ``\textit{pasta}'' is unlikely to be coupled with the aspect category ``\textit{service general}'' due to the large semantic gap between them. But such information cannot be suitably utilized in those classification-type methods.

Inspired by recent success in formulating various NLP tasks as text generation problems \cite{athiwaratkun-etal-2020-augmented, iclr21-augmented, gpt-understands}, we propose to tackle ASQP in a sequence-to-sequence (S2S) manner in this paper.
On one hand, the sentiment quads can be predicted in an end-to-end manner, alleviating the potential error propagation in the pipeline solutions.
On the other hand, the rich label semantic information could be fully exploited by learning to generate the sentiment elements in the natural language form.

Exploiting generation modeling for the ASQP task mainly faces two challenges: 
(i) how to linearize the desired sentiment information so as to facilitate the S2S learning? (ii) how can we utilize the pretrained models for tackling the task, which is a common practice now for solving various ABSA tasks \cite{emnlp20-xulu, coling20-acsa}?
To handle these two challenges, we propose a novel \textsc{Paraphrase} modeling paradigm, which transforms the ASQP task as a paraphrase generation problem \cite{cl13-paraphrase}. 
Specifically, our approach linearizes the sentiment quad into a natural language sentence as if we were paraphrasing the input sentence and highlighting its major sentiment elements.
For example, we can transform the aforementioned sentiment quad (\textit{food quality}, \textit{pasta}, \textit{over-cooked}, \textit{negative}) to a sentence ``\textit{Food quality is bad because pasta is over-cooked}''.
Such a linearized target sequence, paired with the input sentence ``\textit{The pasta is over-cooked!}'' can then be used to learn the mapping function of a generation model. We can seamlessly utilize the large pretrained generative models such as T5 \cite{t5-paper} by fine-tuning with such input-target pairs. Therefore, the rich label semantics of the sentiment elements is naturally fused with the rich knowledge of the pretrained models in the form of natural sentences, rather than directly treating the desired sentiment quad text sequence as the generation target \cite{acl21-gabsa}.

We summarize our contributions as follows:
1) We study a new task, namely aspect sentiment quad prediction (ASQP) in this work and introduce two datasets with sentiment quad annotations for each sample, aiming to analyze more comprehensive aspect-level sentiment information.
2) We propose to tackle ASQP as a paraphrase generation problem, which can predict the sentiment quads in one shot and fully utilize the semantics information of natural language labels.
3) Extensive experiments show that the proposed \textsc{Paraphrase} modeling is effective to tackle ASQP as well as other ABSA tasks, outperforming the previous state-of-the-art models in all cases.
4) The experiment also suggests that our \textsc{Paraphrase} method naturally facilitates the knowledge transfer across related tasks with the unified framework, which can be especially beneficial in the low-resource setting.\footnote{Code and annotated ASQP datasets are publicly available at \url{https://github.com/IsakZhang/ABSA-QUAD}.}

\section{Related Work}
ABSA has been extensively studied in recent years where the main research line is the extraction of the sentiment elements. 
Early studies focus on the prediction of a single element such as extracting the aspect term \cite{emnlp15-ate, ijcai16-ate, acl18-ate-xuhu, acl19-s2s-ate}, detecting the mentioned aspect category \cite{aaai15-acp, naacl21-asap}, and predicting the sentiment polarity, given either an aspect term \cite{emnlp16-asc, emnlp18-asc, emnlp20-asc} or an aspect category \cite{emnlp16-ruder-asc-category, emnlp19-asc-category}.
Some works further consider the joint detection of two sentiment elements, including the pairwise extraction of aspect and opinion term \cite{aaai17-cmla, acl20-aope, acl20-spanmlt}; the prediction of aspect term and its corresponding sentiment polarity \cite{aaai19-lx-e2e-tbsa, acl19-ruidan, acl19-span, acl19-luo-doer, acl20-racl}; and the co-extraction of aspect category and sentiment polarity \cite{coling20-acsa}.

More recently, triplet prediction tasks are proposed in ABSA, aiming to predict the sentiment elements in triplet format. \citet{aaai20-robin} propose the aspect sentiment triplet extraction (ASTE) task, which has received lots of attention \cite{emnlp20-xulu, arxiv-two-stage-aste, aaai21-uabsa, aaai21-mrc-aste-2}. \citet{aaai20-tasd} introduce the target aspect sentiment detection (TASD) task, aiming to predict the aspect category, aspect term, and sentiment polarity simultaneously, which can handle the case where the aspect term is implicit expressed in the given text (treated as ``null'') 
\cite{kbs21-mejd-tasd}.
Built on top of those tasks, we introduce the aspect sentiment quad prediction problem, aiming to predict the four sentiment elements in one shot, which can provide a more detailed and comprehensive sentiment structure for a given text.

Adopting pretrained transformer-based models such as BERT \cite{bert} has become a common practice for tackling the ABSA problem. Especially, many ABSA tasks benefit from appropriately utilizing the pretrained models. \citet{naacl19-sunchi} transform the aspect sentiment classification task as a language inference problem by constructing an auxiliary sentence. \citet{aaai21-mrc-aste-2} and \citet{aaai21-uabsa} formulate multiple ABSA tasks as a reading comprehension task to fully utilize the knowledge of the pre-trained model. Very recently, there are some attempts on tackling ABSA problem in a S2S manner, either treating the class index \cite{acl21-qxp} or the desired sentiment element sequence \cite{acl21-gabsa} as the target of the generation model. In this work, we propose a \textsc{Paraphrase} modeling that can better utilize the knowledge of the pre-trained model via casting the original task to a paraphrase generation process.

\section{Methodology}

\subsection{Problem Statement}
Given a sentence $\boldsymbol{x}$, aspect sentiment quad prediction (ASQP) aims to predict all aspect-level sentiment quadruplets $\{(c, a, o, p)\}$ which corresponds to the aspect category, aspect term, opinion term, and sentiment polarity, respectively. The aspect category $c$ falls into a category set $V_c$; the aspect term $a$ and the opinion term $o$ are typically text spans in the sentence $\boldsymbol{x}$ while the aspect term can also be null if the target is not explicitly mentioned: $a \in V_{\boldsymbol{x}} \cup \{\varnothing\}$ and $o \in V_{\boldsymbol{x}}$ where $V_{\boldsymbol{x}}$ denotes the set containing all possible continuous spans of $\boldsymbol{x}$. The sentiment polarity $p$ belongs to one of the sentiment class \{\texttt{POS}, \texttt{NEU}, \texttt{NEG}\} denoting the positive, neutral, and negative sentiment respectively.

\subsection{ASQP as Paraphrase Generation} \label{sec:paraphrase}
\begin{figure}
    \centering
    \includegraphics[width=\columnwidth]{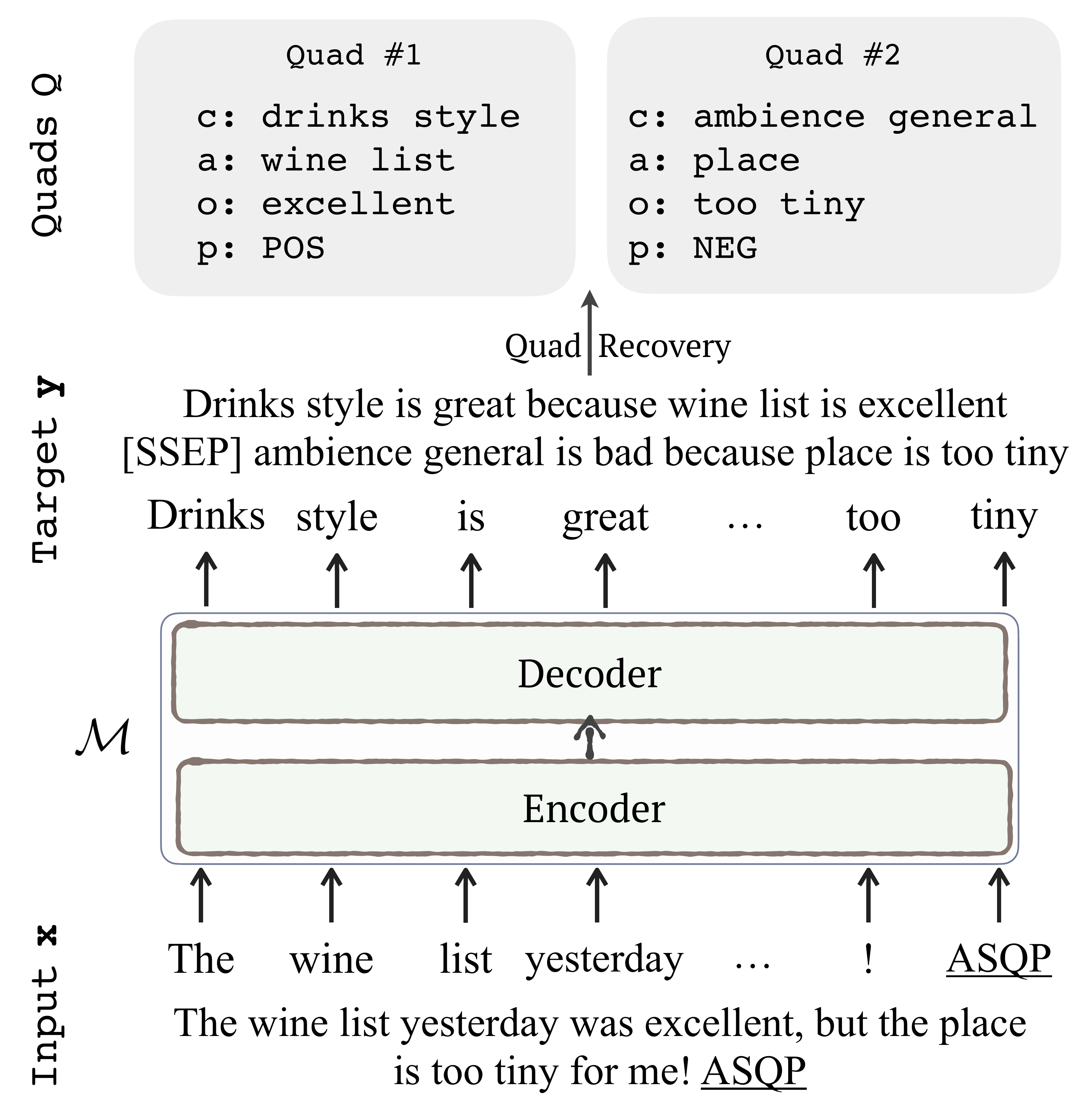}
    \caption{Overview of the paraphrase generation framework. The underlined task identifier in the input is only used under the cross-task transfer setting.}
    \label{fig:model}
    \vspace{-0.4cm}
\end{figure}

We propose a \textsc{Paraphrase} modeling paradigm to transform the ASQP task as a paraphrase generation problem and solve it in a sequence-to-sequence manner. 
As depicted in Figure \ref{fig:model}, given a sentence $\boldsymbol{x}$, we aim to generate a target sequence $\boldsymbol{y}$ with an encoder-decoder model $\mathcal{M}: \boldsymbol{x} \rightarrow \boldsymbol{y}$ where $\boldsymbol{y}$ contains all the desired sentiment elements. Then the sentiment quads $Q=\{(c, a, o, p)\}$ can be recovered from $\boldsymbol{y}$ for making the prediction. 

On one hand, the semantics of the sentiment elements in $Q$ could be fully exploited by generating them in the natural language form in $\boldsymbol{y}$. On the other hand, the input and target are both natural language sentences, which can naturally utilize the rich knowledge in the pretrained generative model. 

\paragraph{\textsc{Paraphrase} Modeling}
To facilitate the S2S learning, given the sentence label pair $(\boldsymbol{x}, Q)$, an important component of the \textsc{Paraphrase} modeling framework is to linearize the sentiment quads $Q$ to a natural language sequence $\boldsymbol{y}$ for constructing the input target pair $(\boldsymbol{x}, \boldsymbol{y})$.

Ideally, we aim to neglect unnecessary details in the input sentence while highlight the major sentiment elements in the target sentence during the paraphrasing process. 
Based on this motivation, we linearize a sentiment quad $q=(c, a, o, p)$ to a natural sentence as follows:
\begin{quote}
    \colorbox{mygray}{
    $\mathcal{P}_c(c)$ is $\mathcal{P}_p(p)$ because $\mathcal{P}_a(a)$ is  $\mathcal{P}_o(o)$.
    } 
\end{quote}
where $\mathcal{P}_z(\cdot)$ is the projection function for $z \in \{c, a, o, p\}$, which maps the sentiment element $z$ from the original format to a natural language form. By adopting suitable projection functions, a structured sentiment quad $q$ can then be transformed to an equivalent natural language sentence.

For the input sentence $\boldsymbol{x}$ with multiple sentiment quads, we first linearize each quad $q$ to a natural sentence as described above. Then these sentences are concatenated with a special symbol \verb|[SSEP]| to form the final target sequence $\boldsymbol{y}$, containing all the sentiment quads for the given sentence. 

\paragraph{Target Construction for ASQP} 
Since the aspect category $c$ and opinion term $o$ in each sentiment quad are already in the natural language form, their projection functions just keep the original formats: $\mathcal{P}_c(c)=c$ and $\mathcal{P}_o(o)=o$. For the sentiment polarity, the projection is as follows:
\begin{equation} \label{opinion-project}
    \mathcal{P}_p(p) =
  \begin{cases}
    \text{\textit{great}}  & \text{if} \; p=\texttt{POS} \\
    \text{\textit{ok}}     & \text{if} \; p=\texttt{NEU} \\
    \text{\textit{bad}}    & \text{if} \; p=\texttt{NEG} \\
  \end{cases} \\
\end{equation}
where the main idea is to transform the sentiment label from the original class format to a natural language expression and also ensure the coherence of the whole linearized target sequence so that the semantics of the sentiment polarity can be exploited by the generation model. Note that the specific mapping can either be pre-defined with commonsense knowledge as in Equation \ref{opinion-project} or dataset-dependent which utilizes the most common concurring opinion term for each sentiment polarity as the sentiment expression.

As for the aspect term, we map it to an implicit pronoun if it is not explicitly mentioned, otherwise we can just use the original natural language form:
\begin{equation}
    \mathcal{P}_a(a) =
  \begin{cases}
    \text{\textit{it}}    & a = \varnothing \\
    a            & otherwise \\
  \end{cases} \\
\end{equation}
This is to mimic the writing process where users often use a pronoun such as ``\textit{it}'' or ``\textit{this}'' to refer to a target that is not explicitly expressed. 

\begin{figure}
    \centering
    \includegraphics[width=\columnwidth]{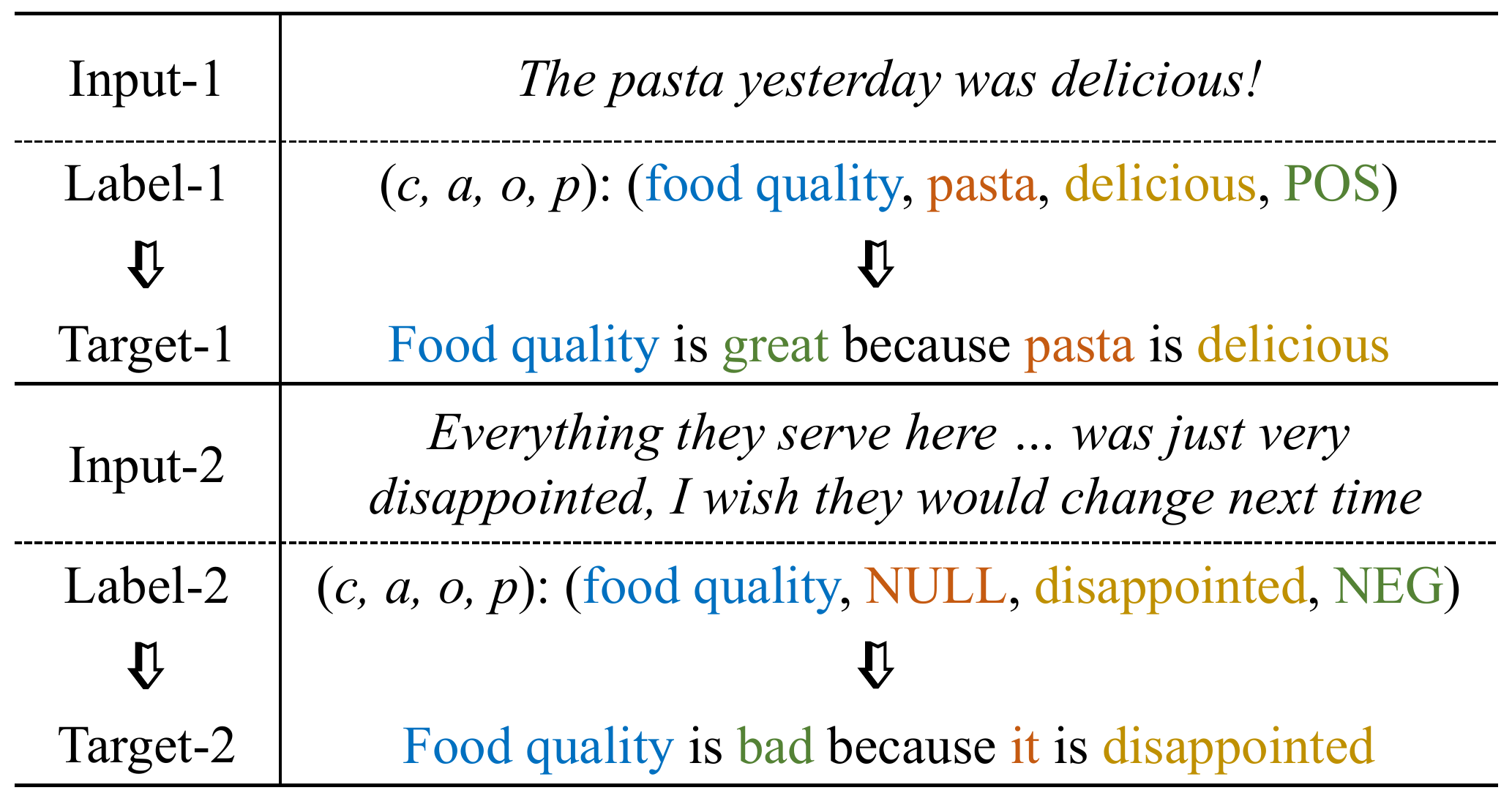}
    \caption{Two examples of the target sentence construction for the ASQP task. Better viewed in colors.}
    \label{fig:paraphrase}
\end{figure}

After defining the specific projection functions for each sentiment element, we can then transform a sentiment quad to a sentence containing all the elements in the natural language form to facilitate the S2S learning. Two target construction examples for the ASQP task are shown in Figure \ref{fig:paraphrase}.

\subsection{Sequence-to-Sequence Learning} \label{sec:seq2seq}
The input-to-target generation can be modeled with a classical encoder-decoder model such as the Transformer architecture \cite{nips17-transformer}. Given the sentence $\boldsymbol{x}$, the encoder first transforms it into a contextualized encoded sequence $\boldsymbol{e}$. The decoder then aims to model the conditional probability distribution of the target sentence $\boldsymbol{y}$ given the encoded input representation: $p_{\theta}(\boldsymbol{y}|\boldsymbol{e})$ which is parameterized by $\theta$. 

At the $i$-th time step, the decoder output $\boldsymbol{y}_i$ is computed based on both the encoded input $\boldsymbol{e}$ and the previous outputs $\boldsymbol{y}_{<i}$: $\boldsymbol{y}_i = f_{dec}(\boldsymbol{e}, \boldsymbol{y}_{<i})$
where $f_{dec}(\cdot)$ denotes the decoder computations. To obtain the probability distribution for the next token, a softmax function is then applied:
\begin{equation}
    p_{\theta}(\boldsymbol{y}_{i+1}|\boldsymbol{e}, \boldsymbol{y}_{<i+1})=\operatorname{softmax}(W^T \boldsymbol{y}_{i})  
\end{equation}
where $W$ maps the prediction $\boldsymbol{y}_i$ to a logit vector, which can then be used to compute the probability distribution over the whole vocabulary set.

\paragraph{Training}
With a pretrained encoder-decoder model such as T5 \cite{t5-paper}, we can initialize $\theta$ with the pretrained parameter weights and further fine-tune the parameters on the input-target pair to maximize the log-likelihood $p_{\theta}(\boldsymbol{y}|\boldsymbol{e})$:
\begin{gather}
    \max _{\theta} \log p_{\theta}(\boldsymbol{y}|\boldsymbol{e}) =\sum\nolimits_{i=1}^{n} \log p_{\theta}(\boldsymbol{y}_{i}|\boldsymbol{e}, \boldsymbol{y}_{<i})
\end{gather}
where $n$ is the length of the target sequence $\boldsymbol{y}$.

\paragraph{Inference and Quad Recovery}
After the training, we generate the target sequence $\boldsymbol{y'}$ in an autoregressive manner and select the token with the highest probability over the vocabulary set as the next token at each time step.
Then we can recover the predicted sentiment quads $Q'$ from the generations. Specifically, we first split the possible multiple quads via detecting the pre-defined separation token \texttt{[SSEP]}. Then for each linearized sentiment quad sequence, we extract the sentiment elements according to the modeling strategy introduced in Sec \ref{sec:paraphrase} and compare them with the gold sentiment quad in $Q$ for the evaluation.
If such decoding fails, for example, the generated sequence violates the defined format, we treat the prediction as null.

\subsection{ABSA as Paraphrase Generation} \label{sec:absa-para}

The proposed \textsc{Paraphrase} modeling in fact provides a general paradigm to tackle the ABSA problem, which transforms the sentiment element prediction to a paraphrase generation process. Therefore, it can be easily extended to handle other ABSA tasks as well: we only need to change the projection functions for each sentiment element to suit the need for each task. We take the target aspect sentiment detection (TASD) \cite{aaai20-tasd} and aspect sentiment triplet extraction (ASTE) \cite{aaai20-robin} tasks as two examples here\footnote{In fact, any ABSA task involving the prediction of one or multiple sentiment elements can be considered as a sub-task of ASQP. We mainly discuss ASTE and TASD tasks in this paper since they are more closely related.}.

The TASD task predicts the $(c, a, p)$ triplets where all sentiment elements have the same condition as in the ASQP problem. Since it does not involve the opinion term prediction, we just let $\mathcal{P}_o(o) = \mathcal{P}_p(p)$ which uses a manually constructed opinion word as the opinion expression to describe the sentiment in the paraphrase. Other projection functions can remain the same as in the ASQP task. For instance, it transforms the (\textit{service general}, \textit{waiter}, \texttt{NEG}) triplet to the target sentence ``\textit{Service general is bad because waiter is bad}''.

For the ASTE task aiming to predict $(a, o, p)$ triplets, we map the aspect category to an implicit pronoun such as ``\textit{it}'' ($\mathcal{P}_o(o)=$ \textit{it}) in all cases.
Besides, it ignores the implicit aspect term, which means $a \in V_{\boldsymbol{x}}$. We then always use the aspect term in its original natural language form: $\mathcal{P}_a(a) = a$. 
Given an example triplet (\textit{Chinese food}, \textit{nice}, \texttt{POS}), a target sentence ``\textit{It is great because Chinese food is nice}'' can be constructed accordingly.

\subsection{Cross-task Knowledge Transfer}
In practice, it is usually rather difficult and expensive to collect large-scale annotated data for complex ABSA problems like ASQP. Fortunately, as introduced in the last section, the proposed \textsc{Paraphrase} method tackles various ABSA tasks in a unified framework.
This characteristic naturally enables the knowledge to be easily transferred across related ABSA tasks, which is especially beneficial under the low-resource setting (\textit{i.e.}, the labeled data for the concerned task is insufficient).

We investigate cross-task transfer for the concerned ASQP task, with the help of its two sub-tasks, including ASTE and TASD.
Similar to recent works on using ``prompt'' as the task identifier \cite{t5-paper, gpt-understands}, we add a task-specific text suffix (\textit{e.g.}, \verb|ASQP| for the ASQP task in Figure \ref{fig:model}) to the input sentence before feeding it to the model for specifying which task the model should perform.
Since the \textsc{Paraphrase} paradigm provides a consistent training objective, the rich task-specific knowledge can first be learned from training on the TASD and ASTE tasks, and then naturally transferred to the ASQP task via fine-tuning on the (limited) ASQP data.

\section{Experimental Setup}

\begin{table}[!t]
    \centering
    \resizebox{\columnwidth}{!}{
    \begin{tabular}{lcccc|cccc}
    \toprule
    & \multicolumn{4}{c}{$\mathtt{Rest15}$} & \multicolumn{4}{c}{$\mathtt{Rest16}$} \\
    & \#S & \#+ & \#0 & \#- & \#S & \#+ & \#0 & \#- \\
    \midrule
    \textbf{Train} & 834 & 1005 & 34 & 315 & 1264 & 1369 & 62 & 558 \\
    \textbf{Dev} & 209 & 252 & 14 & 81 & 316 & 341 & 23 & 143 \\
    \textbf{Test} & 537 & 453 & 37 & 305 & 544 & 583 & 40 & 176 \\
    \bottomrule
    \end{tabular}}
    \caption{Data statistics for the ASQP task. \#S, \#+, \#0,  and \#- denote the number of sentences, number of positive, neutral, negative quads respectively.}
    \label{tab:data}
    \vspace{-0.4cm}
\end{table}

\paragraph{Dataset}

\begin{table*}[!t]
    \centering
    \resizebox{0.9\linewidth}{!}{
    \begin{tabular}{ll|ccc|ccc}
    \toprule
    \multirow{2}{*}{Type} & \multirow{2}{*}{Methods} & \multicolumn{3}{c}{$\mathtt{Rest15}$} & \multicolumn{3}{c}{$\mathtt{Rest16}$} \\
    \cmidrule(lr){3-5} \cmidrule(lr){6-8} 
    & & $\mathtt{Pre}$ & $\mathtt{Rec}$ & $\mathtt{F1}$ & $\mathtt{Pre}$ & $\mathtt{Rec}$ & $\mathtt{F1}$  \\
    \midrule
    \multirow{2}{*}{Pipeline}
    & HGCN-BERT + BERT-Linear & 24.43 &20.25 &22.15 &25.36 &24.03 &24.68\\
    & HGCN-BERT + BERT-TFM & 25.55 &22.01 &23.65 &27.40 &26.41 &26.90\\
    
    \multirow{3}{*}{Unified} 
    &TASO-BERT-Linear   & 41.86 & 26.50 & 32.46 & 49.73 & 40.70 & 44.77 \\
    &TASO-BERT-CRF      & 44.24 & 28.66 & 34.78 & 48.65 & 39.68 & 43.71  \\
    &GAS & \underline{45.31} & \underline{46.70} & \underline{45.98} & \underline{54.54} & \underline{57.62} & \underline{56.04} \\
    \midrule
    \multirow{4}{*}{Ours} 
    &\textsc{Paraphrase} & \textbf{46.16} & \textbf{47.72} & \textbf{46.93} & \textbf{56.63} & \textbf{59.30} & \textbf{57.93} \\
    &\quad w/o sentiment polarity semantics & 45.30 & 46.87 & 46.07 & 56.56 & 58.82 & 57.67  \\
    &\quad w/o aspect category semantics  &  44.65 & 46.59 & 45.60 & 56.27 & 58.38 & 57.31  \\
    &\quad w/o polarity \& category semantics & 43.46 & 45.19 & 44.30 & 56.04 & 57.53 & 56.77 \\
    \bottomrule
    \end{tabular}}
    \caption{Main results of the ASQP task and ablations on label semantics for the proposed method. The best and second best results are in bold and underlined respectively. Scores are averaged over 5 runs with different seeds.}
    \label{tab:asqp}
    \vspace{-0.3cm}
\end{table*}

We build the ASQP datasets based on SemEval Shared Challenges \cite{semeval15-absa, semeval16-absa}. The annotations of the opinion term and aspect category are derived from \citet{aaai20-robin} and \citet{aaai20-tasd} respectively. We align the samples from these two sources and merge the annotations with the same aspect term in each sentence as the anchor. We further conduct some additional annotations:
\begin{itemize}[leftmargin=*]
    \setlength{\itemsep}{0pt}
    \setlength{\parsep}{0pt}
    \setlength{\parskip}{0pt}
    \item Sentences without explicit aspect terms are ignored in \citet{aaai20-robin}, we add these sentences back to our ASQP datasets and manually annotate the opinion terms for them, based on the given aspect category. For example, given a sentence ``\textit{Everything we had was good...}'' with implicit aspect term, we then annotate ``\textit{good}'' as the opinion term according to the aspect category ``\textit{food quality}''. The quads with implicit opinion expressions are discarded.
    
    \item For the same aspect term associated with multiple aspect categories and/or opinion terms, the merging result will have more than four sentiment elements for each quad, we then manually check those cases to correct the labels to ensure the aspect category and opinion term are matched in the same quad.   
    
\end{itemize}

Every sample is annotated by two human annotators and the conflict cases would be checked. Finally, we obtain two datasets, namely $\mathtt{Rest15}$ and $\mathtt{Rest16}$, where each data instance contains a review sentence with one or multiple sentiment quads. We further split 20\% of the data from the training set as the validation set. The statistics is summarized in Table \ref{tab:data}.

\paragraph{Evaluation Metrics} We employ F1 scores as the main evaluation metrics. A sentiment quad prediction is counted as correct if and only if all the predicted elements are exactly the same as the gold labels. We also report the precision (\verb|Pre|) and recall (\verb|Rec|) scores for the ASQP task.

\paragraph{Experiment Details}
The averaged scores over five runs with different random seed initialization are reported.
We adopt the \textsc{T5-Base} \cite{t5-paper} as the pretrained generative model described in Sec \ref{sec:seq2seq}, which adopts a classical Transformer encoder-decoder network architecture.
Regarding the training, we use a batch size of 16 and learning rate being 3e-4. The number of training epochs is 20 for all experiments. 
During the inference, we utilize greedy decoding for generating the output sequence. We also experiment with beam search decoding with the number of beams being 3, 5, and 8 respectively, all leading to similar performance with the greedy decoding. Therefore, greedy decoding is used for simplicity.

\paragraph{Baselines}
Since the ASQP task has not been explored previously, we construct two types of baselines to compare with our \textsc{Paraphrase} method:
\begin{itemize}[leftmargin=*]
    \setlength{\itemsep}{0pt}
    \setlength{\parsep}{0pt}
    \setlength{\parskip}{0pt}
    
    \item \textbf{\textit{Pipeline model}}: we cascade models in a pipeline manner for the quad prediction: \textbf{HGCN} \cite{coling20-acsa} for jointly detecting the aspect category and sentiment polarity, followed by a BERT-based model extracting the aspect and opinion term \cite{wnut19-absa-bert}, given the predicted aspect category and sentiment. The latter one can be either equipped with a linear layer (\textbf{BERT-Linear}) or a transformer block (\textbf{BERT-TFM}) on top.
    
    \item \textbf{\textit{Unified model}}: we first modify TAS \cite{aaai20-tasd}, a state-of-the-art unified model to extract $(c, a, p)$ triplet, for tackling the ASQP task.
    TAS expands each original data sample into multiple samples, each with a specific aspect category and sentiment polarity pair, to solve the task in an end-to-end manner.
    We change its tagging schema to predict aspect and opinion term simultaneously for constructing a unified model to predict the quad, denoted as TASO (TAS with Opinion). There are two variants in terms of the prediction layer: either using a linear classification layer (\textbf{TASO-Linear}) or the CRF layer (\textbf{TASO-CRF}).
    We also consider a generation-type baseline \textbf{GAS}, originally proposed in \cite{acl21-gabsa}, we modify it to directly treat the sentiment quads sequence as the target for learning the generation model. It uses the same pretrained model as ours. 
\end{itemize}

\section{Results and Discussions}

\subsection{Main Results}

The result for the ASQP task is reported in Table \ref{tab:asqp}. There are some notable observations:
Firstly, the performance of the pipeline methods is far from satisfactory. Although both adopting BERT as the backbone, the unified methods (\textit{e.g.}, TASO-BERT-Linear) perform much better than the pipeline ones (\textit{e.g.}, HGCN-BERT + BERT-Linear). 
This verifies our assumption that the pipeline solutions tend to accumulate errors from the sub-task models and finally affect the performance of the final quad prediction.
Secondly, among the unified methods, GAS outperforms two variants of TASO by a large margin, showing the effectiveness of the sequence-to-sequence modeling for the ASQP task. Besides, to solve the task in a unified manner, TASO expands the dataset to $|V_c|\times |V_p|$ times the original size, leading to large computation costs and training time.
Thirdly, we can see that our proposed method, \textsc{Paraphrase} modeling achieves the best performance on all metrics across two datasets. Our method tackles the ASQP problem in an end-to-end manner, alleviating the possible error propagation in the pipeline solutions.
Moreover, compared with the GAS method using the same pre-trained model, our \textsc{Paraphrase} also achieves superior results, suggesting that constructing target sequence in the natural language form is a better way for utilizing the knowledge from the pre-trained generative model, thus leading to better performance.

\subsection{Effect of Label Semantics}
Different from previous classification-type methods for tackling ABSA problem, our \textsc{Paraphrase} modeling can take advantage of the semantics of the sentiment elements by generating the natural language labels. We conduct ablation studies to further investigate the impact of the label semantics. 
Specifically, instead of mapping the label to the natural language form with the projection functions introduced in Sec \ref{sec:paraphrase}, we map each label to a special symbol, similar as the number index in the classification-type models, for representing each label class. 
We consider three cases: (1) w/o sentiment polarity semantics: $\mathcal{P}_p(p_i) =\texttt{SPi}$ where $p_i$ is a sentiment polarity, $i$ denotes the index. For example, we map the positive class as \verb|SP1|;
(2) w/o aspect category semantics: $\mathcal{P}_c(c_j)=\texttt{ACj}$ where we project the aspect category $c_j$ to a symbol with its index $j$\footnote{The mapping relation between the category and their indexes is pre-defined based on the entire dataset.}. For instance, the aspect category ``food quality'' will be mapped to \verb|AC3|;
(3) w/o polarity \& category semantics: it considers the above two cases where both the meaning of aspect category and the sentiment polarity are removed. 

The results are presented in the lower part in Table \ref{tab:asqp}. We can see that discarding the semantics of either element leads to a performance drop, and the drop becomes larger after discarding both of them.
Comparing the ablations on the sentiment polarity and aspect category, the model suffers more when the aspect category is projected to an indexed symbol. The possible reason is that there are only three types of sentiment polarities, which is much less than the number of types for the aspect category. Therefore, it can be easier for the model to learn the mapping between the special symbols and the polarity type during the training.

\begin{table}[!t]
    \centering
    \resizebox{\columnwidth}{!}{
    \begin{tabular}{lcccc}
    \toprule
    & $\mathtt{L14}$ & $\mathtt{R14}$ & $\mathtt{R15}$ & $\mathtt{R16}$ \\
    \midrule
    CMLA+ \small{\cite{aaai17-cmla}}         & 33.16 & 42.79 & 37.01 & 41.72 \\
    Li-unified-R \small{\cite{aaai19-lx-e2e-tbsa}}  & 42.34 & 51.00 & 47.82 & 44.31  \\
    P-pipeline \small{\cite{aaai20-robin}}      & 42.87 & 51.46 & 52.32 & 54.21 \\
    Jet+BERT \small{\cite{emnlp20-xulu}}      & 51.04 & 62.40 & 57.53 & 63.83 \\
    GTS+BERT  \small{\cite{emnlp20-findings-gts}}       & 55.21 & 64.81 & 54.88 & 66.08 \\
    Two-Stage  \small{\cite{arxiv-two-stage-aste}}    & \underline{58.58} & 68.16 & 58.59 & 67.52 \\
    \midrule
    GAS \small{\cite{acl21-gabsa}}  & 58.19 & \underline{70.52} & \underline{60.23} & \underline{69.05} \\
    \textsc{Paraphrase} & \textbf{61.13} & \textbf{72.03} & \textbf{62.56} & \textbf{71.70} \\
    \bottomrule
    \end{tabular}}
    \caption{Results of the ASTE task compared with previous state-of-the-art models. F1 scores are reported.}
    \label{tab:aste}
\end{table}

\subsection{Results on ASTE and TASD Tasks}
As described in Sec \ref{sec:absa-para}, the proposed \textsc{Paraphrase} modeling provides a unified framework to tackle the ABSA problem, we thus test it on the ASTE and TASD tasks, and compare with the previous state-of-the-art methods for each task. 

For the ASTE task, we utilize the dataset provided by \citet{emnlp20-xulu}. We adopt two types of baselines: 1) pipeline-based methods including \textbf{CMLA+} \cite{aaai17-cmla}, \textbf{Li-unified-R} \cite{aaai19-lx-e2e-tbsa}, \textbf{Peng-pipeline} \cite{aaai20-robin} which firstly extract aspect and opinion terms separately, then conduct the pairing; \textbf{Two-stage} \cite{arxiv-two-stage-aste} which proposes a two-stage method to enhance the correlation between aspects and opinions; and 2) end-to-end models including \textbf{GTS} \cite{emnlp20-findings-gts} and \textbf{Jet} \cite{emnlp20-xulu}, both designing unified tagging schemes in order to solve the task in an end-to-end fashion.

For the TASD task, we adopt the dataset prepared by \citet{aaai20-tasd}. We compare with a pipeline-type baseline method \textbf{Baseline-1-f\_lex} \cite{tasd-baseline}, two BERT based models including \textbf{TAS-CRF} and \textbf{TAS-TO} \cite{aaai20-tasd}, and a recent model \textbf{MEJD} \cite{kbs21-mejd-tasd} which utilizes a graph structure to model the dependency among the sentiment elements.

\begin{table}[!t]
    \centering
    \resizebox{0.85\columnwidth}{!}
    {
    \begin{tabular}{lcc}
    \toprule
    & $\mathtt{Rest15}$ & $\mathtt{Rest16}$\\
    \midrule
    \citet{tasd-baseline}  & - & 38.10     \\
    TAS-CRF  \cite{aaai20-tasd}    & 57.51 & 65.89 \\
    TAS-TO   \cite{aaai20-tasd}    & 58.09 & 65.44 \\
    MEJD    \cite{kbs21-mejd-tasd}      & 57.76 & 67.66 \\
    \midrule
    GAS & \underline{60.63} & \underline{68.31} \\
    \textsc{Paraphrase} & \textbf{63.06} & \textbf{71.97} \\
    \bottomrule
    \end{tabular}}
    \caption{Results of the TASD task compared with previous state-of-the-art models. F1 scores are reported.} 
    \label{tab:tasd}
\end{table}

\begin{figure*}
     \centering
     \begin{subfigure}[b]{0.35\textwidth}
         \centering
         \includegraphics[width=\linewidth]{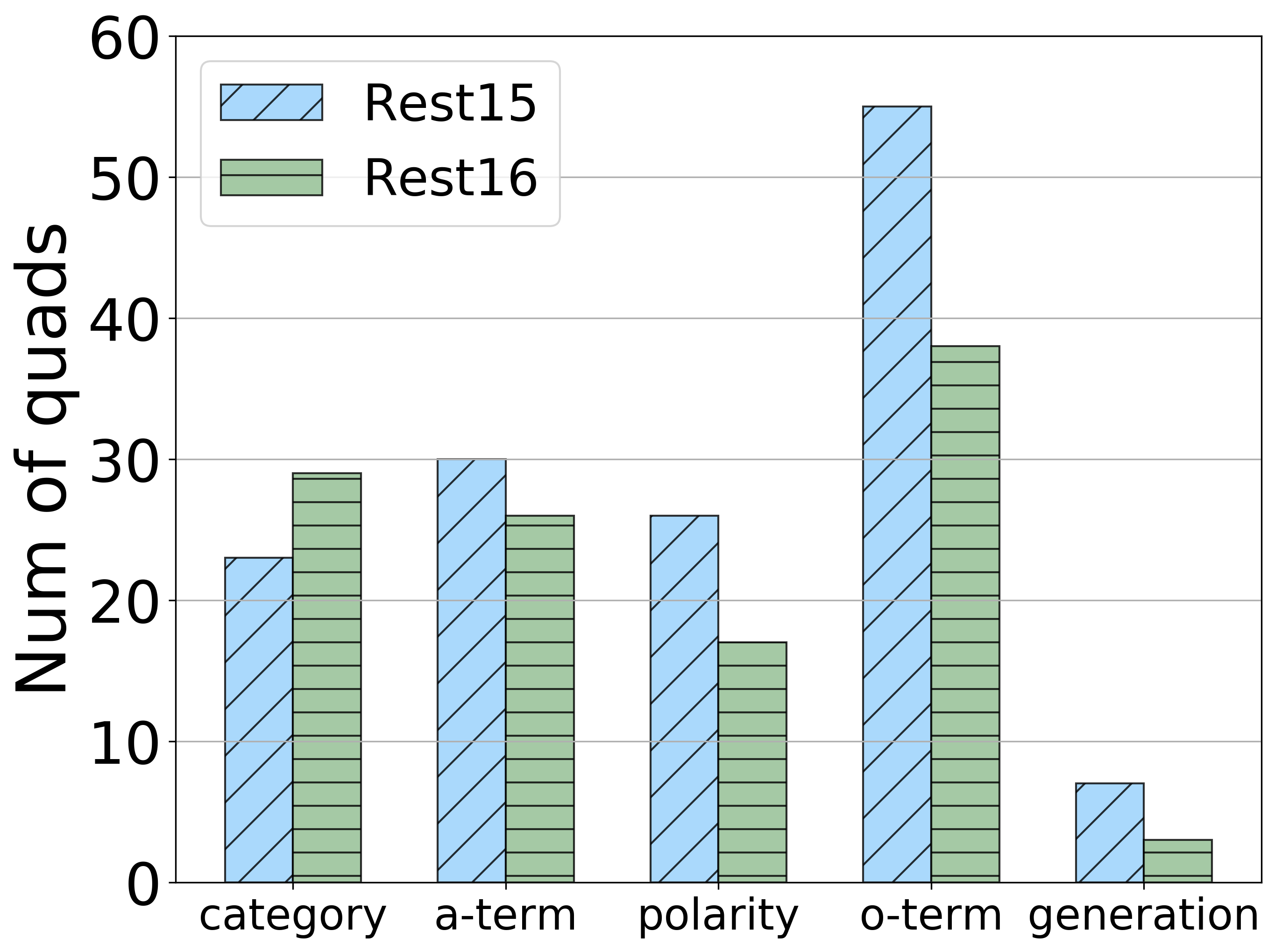}
         \caption{Number of quads w.r.t. the mistake type.}
         \label{error-sentiment-type}
     \end{subfigure}%
     \begin{subfigure}[b]{0.65\textwidth}
         \centering
         \includegraphics[width=\linewidth]{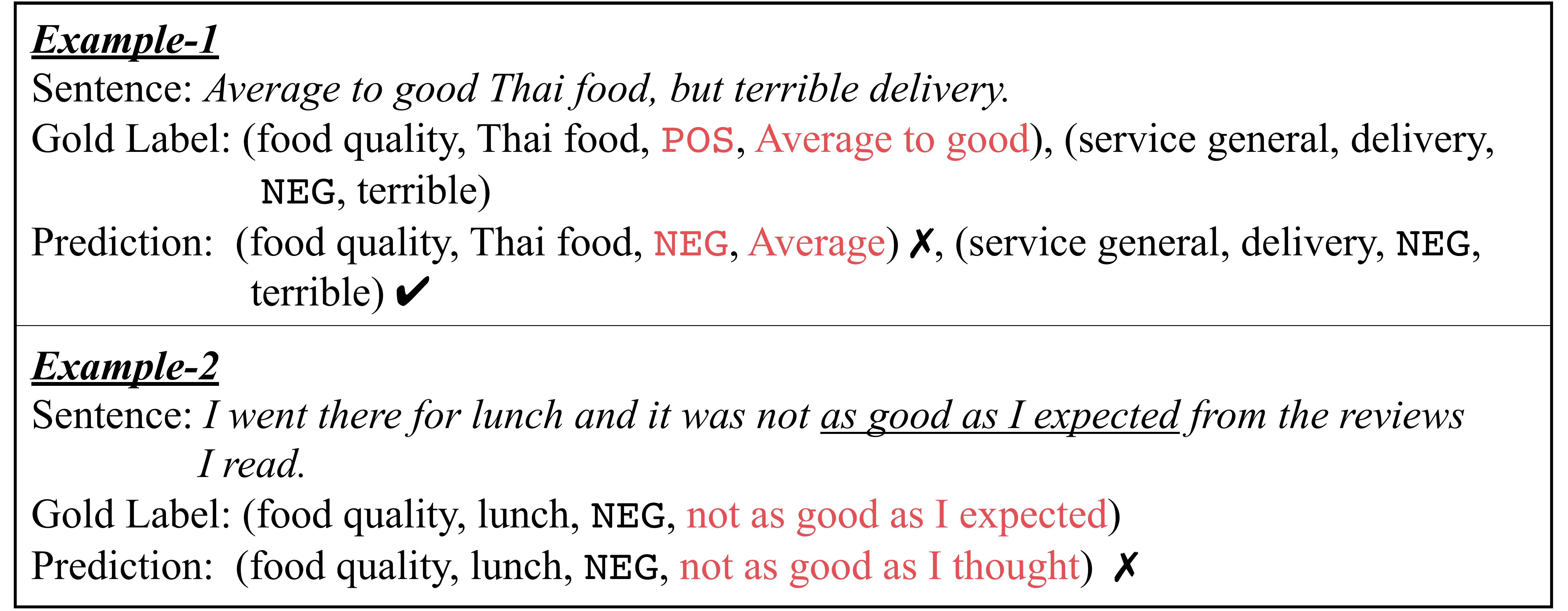}
         \caption{Examples containing the input sentence, gold label and predicted quads.}
         \label{case-study}
     \end{subfigure}
     \caption{Error analysis and case study.}
     \label{graph-relations}
     \vspace{-0.3cm}
\end{figure*}

The results for the ASTE and TASD tasks are shown in the Table \ref{tab:aste} and \ref{tab:tasd} respectively. We also report the performance of the GAS method for comparison. We observe that the proposed \textsc{Paraphrase} method consistently outperforms the previous state-of-the-art models across all datasets in two tasks, showing the effectiveness of converting various ABSA tasks into a paraphrase generation problem. 
More importantly, by transforming the problem into a unified S2S task, we alleviate extensive task-specific model designs. Unlike previous studies with different network architectures for different tasks, we use the same framework for solving the ASQP, ASTE, and TASD tasks, indicating the great generality of the \textsc{Paraphrase} method.

\subsection{Error Analysis and Case Study}
To better understand the behaviour of the proposed method, especially in which cases it would fail,
we conduct error analysis and case study in this section. We sample 100 sentences in the development set of each dataset and employ the trained model to make the predictions. Then we check the incorrect quad predictions and categorize their error types.

We first analyze which type of sentiment element in the sentiment quad is the most difficult for the model to predict and present the results in Figure \ref{error-sentiment-type}. In both datasets, the most common mistake is when predicting the opinion term. Different from the aspect term, opinion term is typically not a single word, but a text span. We find that the model often struggles to detect the exact same span as the ground-truths, as shown in the \underline{\textit{Example-1}} in Figure \ref{case-study}.
For the aspect category, the model is often confused by semantically similar aspect categories such as ``\textit{food quality}'' and ``\textit{food style options}''.
For the sentiment polarity, the most common mistake is made by the confusion between ``positive'' and ``neutral'' classes, possibly due to the imbalanced label distribution in the dataset.

Moreover, we compute the amount of predicted quads whose sentiment elements do not belong to the corresponding vocabulary set, due to the nature of the generation modeling since it does not perform ``extraction'' in the given sentence. For instance, a predicted aspect category does not belong to the defined aspect category set $V_c$. 
As shown in the $\mathtt{generation}$ column in Figure \ref{error-sentiment-type}, this error type in fact accounts for only a small portion in total. \underline{\textit{Example-2}} presents a case for such error where the model changes the word ``\textit{expected}'' in the original sentence to ``\textit{thought}'' when predicting the opinion term. Although this might be similar to human readers, this prediction is judged as incorrect since we use the exact match for the evaluation.
Nevertheless, contrary to the possible perception that the generation type method might generate unbounded contents which can be difficult to recover sentiment quads or provide meaningless outputs, the predictions from the proposed method actually suffer little from the generation error.

\subsection{ABSA Cross-task Transfer}
\begin{figure}
    \centering
    \includegraphics[width=\linewidth]{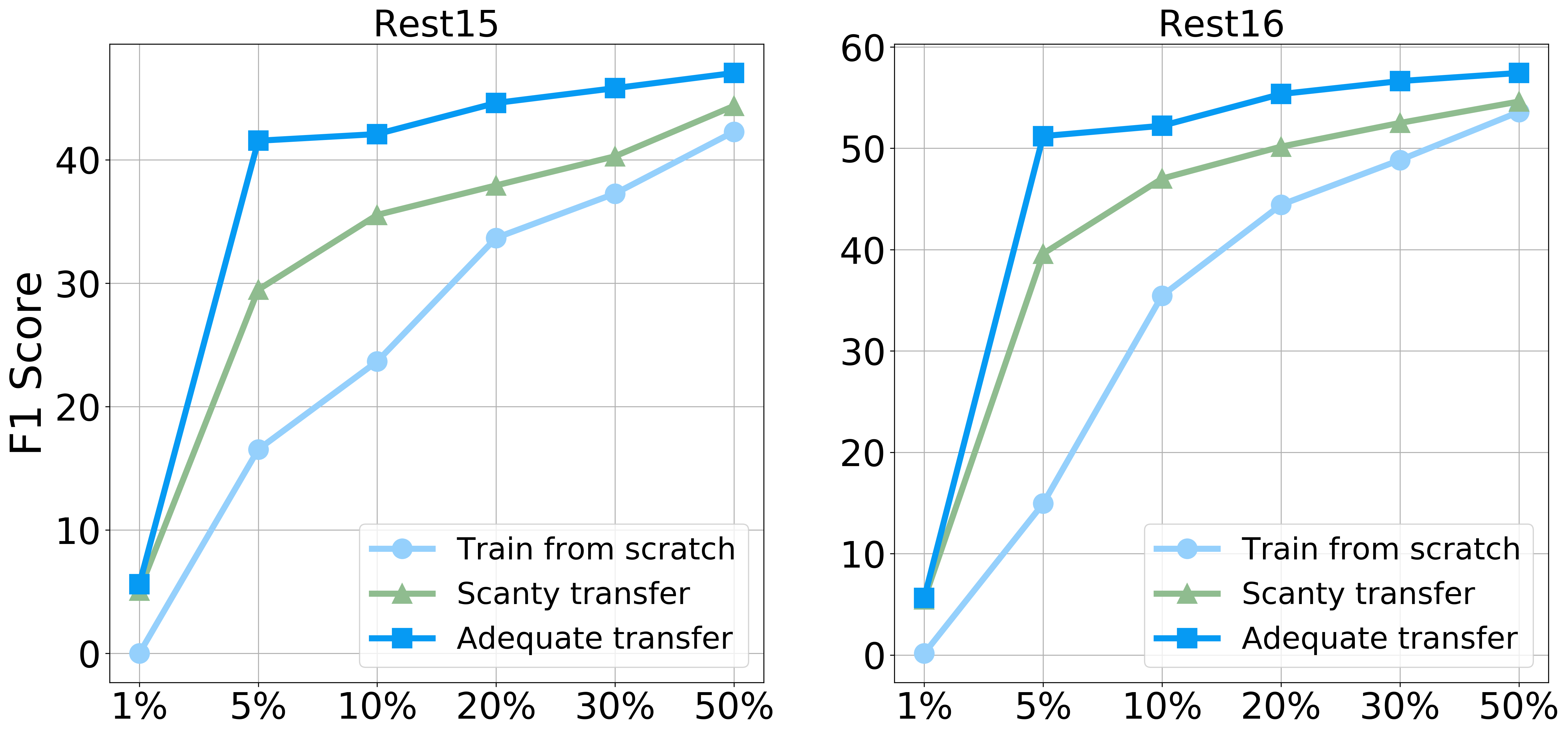}
    \caption{Cross-task transfer results. F1 scores on two datasets are shown with respect to the ratio of the ASQP data under three settings.}
    \label{fig:transfer}
    \vspace{-0.3cm}
\end{figure}

With the \textsc{Paraphrase} modeling, different ABSA tasks can be tackled in a similar manner, enabling the knowledge learned from related tasks to be easily transferred to the target task.
In our case, ASTE and TASD are regarded as two sub-tasks to transfer the knowledge for handling ASQP. 
Here we consider two common situations where we might have adequate ASTE/TASD data for transfer (``Adequate transfer'') or we only have a small amount of ASTE/TASD data (``Scanty transfer''). In the experiment, we utilize 500/100 ASTE and TASD data samples for these two settings respectively.
We vary the ratio of the ASQP data to simulate different scales of low-resource settings and report the results under two transfer situations in Figure \ref{fig:transfer}. We also show the performance if we only train the model with the ASQP task, without any help from the knowledge transfer (``Train from scratch''). 

As can be observed in the figure, utilizing the knowledge learned from two triplet detection tasks can greatly benefit the concerned sentiment quad prediction.
For instance, with adequate annotated data of ASTE and TASD, using 5\% of the ASQP data can lead to competitive performance compared with purely training with 50\% ASQP data.
Even with a scanty amount of data from related tasks to transfer, the model can still perform much better than purely trained on the sentiment quad data, especially under the low-resource setting.

\section{Conclusions}
We introduce a new ABSA task, namely aspect sentiment quad prediction (ASQP) in this paper, aiming to provide a more comprehensive aspect-level sentiment picture. We propose a novel \textsc{Paraphrase} modeling paradigm that tackles the original quad prediction as a paraphrase generation problem. Experiments on two datasets show its superiority compared with previous state-of-the-art models. 
We also demonstrate that the proposed method provides a unified framework that can be easily adapted to handle other ABSA tasks as well. Extensive analysis are conducted to understand the characteristics of the proposed method.

We can notice that ASQP remains a challenging problem
and worth further exploring. We look forward future work could propose better methods to tackle such a difficult ABSA task for fully revealing the aspect-level opinion information.

\typeout{}
\bibliography{custom}

\begin{thebibliography}{44}
\expandafter\ifx\csname natexlab\endcsname\relax\def\natexlab#1{#1}\fi

\bibitem[{Athiwaratkun et~al.(2020)Athiwaratkun, Nogueira~dos Santos, Krone,
  and Xiang}]{athiwaratkun-etal-2020-augmented}
Ben Athiwaratkun, Cicero Nogueira~dos Santos, Jason Krone, and Bing Xiang.
  2020.
\newblock \href {https://doi.org/10.18653/v1/2020.emnlp-main.27} {Augmented
  natural language for generative sequence labeling}.
\newblock In \emph{Proceedings of the 2020 Conference on Empirical Methods in
  Natural Language Processing (EMNLP)}, pages 375--385, Online. Association for
  Computational Linguistics.

\bibitem[{Bhagat and Hovy(2013)}]{cl13-paraphrase}
Rahul Bhagat and Eduard~H. Hovy. 2013.
\newblock \href {https://doi.org/10.1162/COLI\_a\_00166} {What is a
  paraphrase?}
\newblock \emph{Comput. Linguistics}, 39(3):463--472.

\bibitem[{Brun and Nikoulina(2018)}]{tasd-baseline}
Caroline Brun and Vassilina Nikoulina. 2018.
\newblock \href {https://doi.org/10.18653/v1/w18-6217} {Aspect based sentiment
  analysis into the wild}.
\newblock In \emph{Proceedings of the 9th Workshop on Computational Approaches
  to Subjectivity, Sentiment and Social Media Analysis, WASSA@EMNLP 2018},
  pages 116--122.

\bibitem[{Bu et~al.(2021)Bu, Ren, Zheng, Yang, Wang, Zhang, and
  Wu}]{naacl21-asap}
Jiahao Bu, Lei Ren, Shuang Zheng, Yang Yang, Jingang Wang, Fuzheng Zhang, and
  Wei Wu. 2021.
\newblock \href {https://arxiv.org/abs/2103.06605} {{ASAP:} {A} chinese review
  dataset towards aspect category sentiment analysis and rating prediction}.
\newblock \emph{CoRR}, abs/2103.06605.

\bibitem[{Cai et~al.(2020)Cai, Tu, Zhou, Yu, and Xia}]{coling20-acsa}
Hongjie Cai, Yaofeng Tu, Xiangsheng Zhou, Jianfei Yu, and Rui Xia. 2020.
\newblock \href {https://doi.org/10.18653/v1/2020.coling-main.72}
  {Aspect-category based sentiment analysis with hierarchical graph
  convolutional network}.
\newblock In \emph{Proceedings of the 28th International Conference on
  Computational Linguistics, {COLING} 2020}, pages 833--843.

\bibitem[{Chen et~al.(2020)Chen, Liu, Wang, Zhang, and Chi}]{acl20-aope}
Shaowei Chen, Jie Liu, Yu~Wang, Wenzheng Zhang, and Ziming Chi. 2020.
\newblock \href {https://doi.org/10.18653/v1/2020.acl-main.582} {Synchronous
  double-channel recurrent network for aspect-opinion pair extraction}.
\newblock In \emph{Proceedings of the 58th Annual Meeting of the Association
  for Computational Linguistics, {ACL} 2020}, pages 6515--6524.

\bibitem[{Chen et~al.(2021)Chen, Wang, Liu, and Wang}]{aaai21-mrc-aste-2}
Shaowei Chen, Yu~Wang, Jie Liu, and Yuelin Wang. 2021.
\newblock \href {https://arxiv.org/abs/2103.07665} {Bidirectional machine
  reading comprehension for aspect sentiment triplet extraction}.
\newblock In \emph{AAAI}.

\bibitem[{Chen and Qian(2020)}]{acl20-racl}
Zhuang Chen and Tieyun Qian. 2020.
\newblock \href {https://www.aclweb.org/anthology/2020.acl-main.340/}
  {Relation-aware collaborative learning for unified aspect-based sentiment
  analysis}.
\newblock In \emph{Proceedings of the 58th Annual Meeting of the Association
  for Computational Linguistics, {ACL} 2020}, pages 3685--3694.

\bibitem[{Devlin et~al.(2019)Devlin, Chang, Lee, and Toutanova}]{bert}
Jacob Devlin, Ming-Wei Chang, Kenton Lee, and Kristina Toutanova. 2019.
\newblock \href {https://www.aclweb.org/anthology/N19-1423} {{BERT}:
  Pre-training of deep bidirectional transformers for language understanding}.
\newblock In \emph{NAACL}, pages 4171--4186.

\bibitem[{He et~al.(2019)He, Lee, Ng, and Dahlmeier}]{acl19-ruidan}
Ruidan He, Wee~Sun Lee, Hwee~Tou Ng, and Daniel Dahlmeier. 2019.
\newblock \href {https://doi.org/10.18653/v1/p19-1048} {An interactive
  multi-task learning network for end-to-end aspect-based sentiment analysis}.
\newblock In \emph{ACL19}, pages 504--515.

\bibitem[{Hu et~al.(2019{\natexlab{a}})Hu, Zhao, Zhang, Cai, Su, Cheng, and
  Shen}]{emnlp19-asc-category}
Mengting Hu, Shiwan Zhao, Li~Zhang, Keke Cai, Zhong Su, Renhong Cheng, and
  Xiaowei Shen. 2019{\natexlab{a}}.
\newblock \href {https://doi.org/10.18653/v1/D19-1467} {{CAN:} constrained
  attention networks for multi-aspect sentiment analysis}.
\newblock In \emph{Proceedings of the 2019 Conference on Empirical Methods in
  Natural Language Processing and the 9th International Joint Conference on
  Natural Language Processing, {EMNLP-IJCNLP} 2019}, pages 4600--4609.

\bibitem[{Hu et~al.(2019{\natexlab{b}})Hu, Peng, Huang, Li, and
  Lv}]{acl19-span}
Minghao Hu, Yuxing Peng, Zhen Huang, Dongsheng Li, and Yiwei Lv.
  2019{\natexlab{b}}.
\newblock \href {https://doi.org/10.18653/v1/p19-1051} {Open-domain targeted
  sentiment analysis via span-based extraction and classification}.
\newblock In \emph{Proceedings of the 57th Conference of the Association for
  Computational Linguistics, {ACL} 2019}, pages 537--546.

\bibitem[{Huang and Carley(2018)}]{emnlp18-asc}
Binxuan Huang and Kathleen~M. Carley. 2018.
\newblock \href {https://doi.org/10.18653/v1/d18-1136} {Parameterized
  convolutional neural networks for aspect level sentiment classification}.
\newblock In \emph{Proceedings of the 2018 Conference on Empirical Methods in
  Natural Language Processing}, pages 1091--1096.

\bibitem[{Huang et~al.(2021)Huang, Wang, Li, Liu, Zhang, Cheng, Yin, and
  Wang}]{arxiv-two-stage-aste}
Lianzhe Huang, Peiyi Wang, Sujian Li, Tianyu Liu, Xiaodong Zhang, Zhicong
  Cheng, Dawei Yin, and Houfeng Wang. 2021.
\newblock \href {https://arxiv.org/abs/2102.08549} {First target and opinion
  then polarity: Enhancing target-opinion correlation for aspect sentiment
  triplet extraction}.
\newblock \emph{CoRR}, abs/2102.08549.

\bibitem[{Li et~al.(2019{\natexlab{a}})Li, Bing, Li, and
  Lam}]{aaai19-lx-e2e-tbsa}
Xin Li, Lidong Bing, Piji Li, and Wai Lam. 2019{\natexlab{a}}.
\newblock \href {https://doi.org/10.1609/aaai.v33i01.33016714} {A unified model
  for opinion target extraction and target sentiment prediction}.
\newblock In \emph{AAAI}, pages 6714--6721.

\bibitem[{Li et~al.(2019{\natexlab{b}})Li, Bing, Zhang, and
  Lam}]{wnut19-absa-bert}
Xin Li, Lidong Bing, Wenxuan Zhang, and Wai Lam. 2019{\natexlab{b}}.
\newblock \href {https://doi.org/10.18653/v1/D19-5505} {Exploiting {BERT} for
  end-to-end aspect-based sentiment analysis}.
\newblock In \emph{Proceedings of the 5th Workshop on Noisy User-generated
  Text, W-NUT@EMNLP 2019}, pages 34--41.

\bibitem[{Liu(2012)}]{liu-2012-absa}
Bing Liu. 2012.
\newblock \href {https://doi.org/10.2200/S00416ED1V01Y201204HLT016}
  {\emph{Sentiment Analysis and Opinion Mining}}.
\newblock Synthesis Lectures on Human Language Technologies.

\bibitem[{Liu et~al.(2015)Liu, Joty, and Meng}]{emnlp15-ate}
Pengfei Liu, Shafiq~R. Joty, and Helen~M. Meng. 2015.
\newblock \href {https://doi.org/10.18653/v1/d15-1168} {Fine-grained opinion
  mining with recurrent neural networks and word embeddings}.
\newblock In \emph{Proceedings of the 2015 Conference on Empirical Methods in
  Natural Language Processing, {EMNLP} 2015}, pages 1433--1443.

\bibitem[{Liu et~al.(2021)Liu, Zheng, Du, Ding, Qian, Yang, and
  Tang}]{gpt-understands}
Xiao Liu, Yanan Zheng, Zhengxiao Du, Ming Ding, Yujie Qian, Zhilin Yang, and
  Jie Tang. 2021.
\newblock \href {https://arxiv.org/abs/2103.10385} {{GPT} understands, too}.
\newblock \emph{CoRR}, abs/2103.10385.

\bibitem[{Luo et~al.(2019)Luo, Li, Liu, and Zhang}]{acl19-luo-doer}
Huaishao Luo, Tianrui Li, Bing Liu, and Junbo Zhang. 2019.
\newblock \href {https://doi.org/10.18653/v1/p19-1056} {{DOER:} dual
  cross-shared {RNN} for aspect term-polarity co-extraction}.
\newblock In \emph{ACL}, pages 591--601.

\bibitem[{Ma et~al.(2019)Ma, Li, Wu, Xie, and Wang}]{acl19-s2s-ate}
Dehong Ma, Sujian Li, Fangzhao Wu, Xing Xie, and Houfeng Wang. 2019.
\newblock \href {https://doi.org/10.18653/v1/p19-1344} {Exploring
  sequence-to-sequence learning in aspect term extraction}.
\newblock In \emph{Proceedings of the 57th Conference of the Association for
  Computational Linguistics, {ACL} 2019}, pages 3538--3547.

\bibitem[{Mao et~al.(2021)Mao, Shen, Yu, and Cai}]{aaai21-uabsa}
Yue Mao, Yi~Shen, Chao Yu, and Longjun Cai. 2021.
\newblock \href {https://arxiv.org/abs/2101.00816} {A joint training dual-mrc
  framework for aspect based sentiment analysis}.
\newblock \emph{CoRR}, abs/2101.00816.

\bibitem[{Paolini et~al.(2021)Paolini, Athiwaratkun, Krone, Ma, Achille,
  ANUBHAI, dos Santos, Xiang, and Soatto}]{iclr21-augmented}
Giovanni Paolini, Ben Athiwaratkun, Jason Krone, Jie Ma, Alessandro Achille,
  RISHITA ANUBHAI, Cicero~Nogueira dos Santos, Bing Xiang, and Stefano Soatto.
  2021.
\newblock \href {https://openreview.net/forum?id=US-TP-xnXI} {Structured
  prediction as translation between augmented natural languages}.
\newblock In \emph{International Conference on Learning Representations}.

\bibitem[{Peng et~al.(2020)Peng, Xu, Bing, Huang, Lu, and Si}]{aaai20-robin}
Haiyun Peng, Lu~Xu, Lidong Bing, Fei Huang, Wei Lu, and Luo Si. 2020.
\newblock \href {https://aaai.org/ojs/index.php/AAAI/article/view/6383}
  {Knowing what, how and why: {A} near complete solution for aspect-based
  sentiment analysis}.
\newblock In \emph{The Thirty-Fourth {AAAI} Conference on Artificial
  Intelligence, {AAAI} 2020}, pages 8600--8607.

\bibitem[{Pontiki et~al.(2016)Pontiki, Galanis, Papageorgiou, Androutsopoulos,
  Manandhar, Al{-}Smadi, Al{-}Ayyoub, Zhao, Qin, Clercq, Hoste, Apidianaki,
  Tannier, Loukachevitch, Kotelnikov, Bel, Zafra, and Eryigit}]{semeval16-absa}
Maria Pontiki, Dimitris Galanis, Haris Papageorgiou, Ion Androutsopoulos,
  Suresh Manandhar, Mohammad Al{-}Smadi, Mahmoud Al{-}Ayyoub, Yanyan Zhao, Bing
  Qin, Orph{\'{e}}e~De Clercq, V{\'{e}}ronique Hoste, Marianna Apidianaki,
  Xavier Tannier, Natalia~V. Loukachevitch, Evgeniy~V. Kotelnikov, N{\'{u}}ria
  Bel, Salud Mar{\'{\i}}a~Jim{\'{e}}nez Zafra, and G{\"{u}}lsen Eryigit. 2016.
\newblock \href {https://doi.org/10.18653/v1/s16-1002} {Semeval-2016 task 5:
  Aspect based sentiment analysis}.
\newblock In \emph{Proceedings of the 10th International Workshop on Semantic
  Evaluation, SemEval@NAACL-HLT 2016}, pages 19--30.

\bibitem[{Pontiki et~al.(2015)Pontiki, Galanis, Papageorgiou, Manandhar, and
  Androutsopoulos}]{semeval15-absa}
Maria Pontiki, Dimitris Galanis, Haris Papageorgiou, Suresh Manandhar, and Ion
  Androutsopoulos. 2015.
\newblock \href {https://doi.org/10.18653/v1/s15-2082} {Semeval-2015 task 12:
  Aspect based sentiment analysis}.
\newblock In \emph{SemEval@NAACL-HLT}, pages 486--495.

\bibitem[{Pontiki et~al.(2014)Pontiki, Galanis, Pavlopoulos, Papageorgiou,
  Androutsopoulos, and Manandhar}]{semeval14-absa}
Maria Pontiki, Dimitris Galanis, John Pavlopoulos, Harris Papageorgiou, Ion
  Androutsopoulos, and Suresh Manandhar. 2014.
\newblock \href {https://doi.org/10.3115/v1/s14-2004} {Semeval-2014 task 4:
  Aspect based sentiment analysis}.
\newblock In \emph{SemEval@COLING 2014}, pages 27--35.

\bibitem[{Raffel et~al.(2020)Raffel, Shazeer, Roberts, Lee, Narang, Matena,
  Zhou, Li, and Liu}]{t5-paper}
Colin Raffel, Noam Shazeer, Adam Roberts, Katherine Lee, Sharan Narang, Michael
  Matena, Yanqi Zhou, Wei Li, and Peter~J. Liu. 2020.
\newblock \href {http://jmlr.org/papers/v21/20-074.html} {Exploring the limits
  of transfer learning with a unified text-to-text transformer}.
\newblock \emph{J. Mach. Learn. Res.}, 21:140:1--140:67.

\bibitem[{Ruder et~al.(2016)Ruder, Ghaffari, and
  Breslin}]{emnlp16-ruder-asc-category}
Sebastian Ruder, Parsa Ghaffari, and John~G. Breslin. 2016.
\newblock \href {https://doi.org/10.18653/v1/d16-1103} {A hierarchical model of
  reviews for aspect-based sentiment analysis}.
\newblock In \emph{Proceedings of the 2016 Conference on Empirical Methods in
  Natural Language Processing, {EMNLP} 2016}, pages 999--1005.

\bibitem[{Sun et~al.(2019)Sun, Huang, and Qiu}]{naacl19-sunchi}
Chi Sun, Luyao Huang, and Xipeng Qiu. 2019.
\newblock \href {https://doi.org/10.18653/v1/n19-1035} {Utilizing {BERT} for
  aspect-based sentiment analysis via constructing auxiliary sentence}.
\newblock In \emph{Proceedings of the 2019 Conference of the North American
  Chapter of the Association for Computational Linguistics: Human Language
  Technologies, {NAACL-HLT} 2019}, pages 380--385.

\bibitem[{Vaswani et~al.(2017)Vaswani, Shazeer, Parmar, Uszkoreit, Jones,
  Gomez, Kaiser, and Polosukhin}]{nips17-transformer}
Ashish Vaswani, Noam Shazeer, Niki Parmar, Jakob Uszkoreit, Llion Jones,
  Aidan~N. Gomez, Lukasz Kaiser, and Illia Polosukhin. 2017.
\newblock \href
  {https://proceedings.neurips.cc/paper/2017/hash/3f5ee243547dee91fbd053c1c4a845aa-Abstract.html}
  {Attention is all you need}.
\newblock In \emph{Advances in Neural Information Processing Systems 30: Annual
  Conference on Neural Information Processing Systems 2017}, pages 5998--6008.

\bibitem[{Wan et~al.(2020)Wan, Yang, Du, Liu, Qi, and Pan}]{aaai20-tasd}
Hai Wan, Yufei Yang, Jianfeng Du, Yanan Liu, Kunxun Qi, and Jeff~Z. Pan. 2020.
\newblock \href {https://aaai.org/ojs/index.php/AAAI/article/view/6447}
  {Target-aspect-sentiment joint detection for aspect-based sentiment
  analysis}.
\newblock In \emph{The Thirty-Fourth {AAAI} Conference on Artificial
  Intelligence, {AAAI} 2020}, pages 9122--9129.

\bibitem[{Wang et~al.(2017)Wang, Pan, Dahlmeier, and Xiao}]{aaai17-cmla}
Wenya Wang, Sinno~Jialin Pan, Daniel Dahlmeier, and Xiaokui Xiao. 2017.
\newblock \href {http://aaai.org/ocs/index.php/AAAI/AAAI17/paper/view/14441}
  {Coupled multi-layer attentions for co-extraction of aspect and opinion
  terms}.
\newblock In \emph{Proceedings of the Thirty-First {AAAI} Conference on
  Artificial Intelligence}, pages 3316--3322.

\bibitem[{Wang et~al.(2016)Wang, Huang, Zhu, and Zhao}]{emnlp16-asc}
Yequan Wang, Minlie Huang, Xiaoyan Zhu, and Li~Zhao. 2016.
\newblock \href {https://doi.org/10.18653/v1/d16-1058} {Attention-based {LSTM}
  for aspect-level sentiment classification}.
\newblock In \emph{Proceedings of the 2016 Conference on Empirical Methods in
  Natural Language Processing, {EMNLP} 2016}, pages 606--615.

\bibitem[{Wu et~al.(2021)Wu, Xiong, Yi, Yu, Zhu, Gao, and
  Chen}]{kbs21-mejd-tasd}
Chao Wu, Qingyu Xiong, Hualing Yi, Yang Yu, Qiwu Zhu, Min Gao, and Jie Chen.
  2021.
\newblock \href
  {https://www.sciencedirect.com/science/article/pii/S0950705121003361}
  {Multiple-element joint detection for aspect-based sentiment analysis}.
\newblock \emph{Knowledge-Based Systems}, page 107073.

\bibitem[{Wu et~al.(2020)Wu, Ying, Zhao, Fan, Dai, and
  Xia}]{emnlp20-findings-gts}
Zhen Wu, Chengcan Ying, Fei Zhao, Zhifang Fan, Xinyu Dai, and Rui Xia. 2020.
\newblock \href {https://www.aclweb.org/anthology/2020.findings-emnlp.234}
  {Grid tagging scheme for aspect-oriented fine-grained opinion extraction}.
\newblock In \emph{Findings of the Association for Computational Linguistics:
  EMNLP 2020}.

\bibitem[{Xu et~al.(2018)Xu, Liu, Shu, and Yu}]{acl18-ate-xuhu}
Hu~Xu, Bing Liu, Lei Shu, and Philip~S. Yu. 2018.
\newblock \href {https://www.aclweb.org/anthology/P18-2094/} {Double embeddings
  and cnn-based sequence labeling for aspect extraction}.
\newblock In \emph{Proceedings of the 56th Annual Meeting of the Association
  for Computational Linguistics, {ACL} 2018}, pages 592--598.

\bibitem[{Xu et~al.(2020)Xu, Li, Lu, and Bing}]{emnlp20-xulu}
Lu~Xu, Hao Li, Wei Lu, and Lidong Bing. 2020.
\newblock \href {https://doi.org/10.18653/v1/2020.emnlp-main.183}
  {Position-aware tagging for aspect sentiment triplet extraction}.
\newblock In \emph{Proceedings of the 2020 Conference on Empirical Methods in
  Natural Language Processing, {EMNLP} 2020}, pages 2339--2349.

\bibitem[{Yan et~al.(2021)Yan, Dai, Ji, Qiu, and Zhang}]{acl21-qxp}
Hang Yan, Junqi Dai, Tuo Ji, Xipeng Qiu, and Zheng Zhang. 2021.
\newblock \href {https://doi.org/10.18653/v1/2021.acl-long.188} {A unified
  generative framework for aspect-based sentiment analysis}.
\newblock In \emph{Proceedings of the 59th Annual Meeting of the Association
  for Computational Linguistics and the 11th International Joint Conference on
  Natural Language Processing, {ACL/IJCNLP} 2021}, pages 2416--2429.

\bibitem[{Yin et~al.(2016)Yin, Wei, Dong, Xu, Zhang, and Zhou}]{ijcai16-ate}
Yichun Yin, Furu Wei, Li~Dong, Kaimeng Xu, Ming Zhang, and Ming Zhou. 2016.
\newblock \href {http://www.ijcai.org/Abstract/16/423} {Unsupervised word and
  dependency path embeddings for aspect term extraction}.
\newblock In \emph{Proceedings of the Twenty-Fifth International Joint
  Conference on Artificial Intelligence, {IJCAI} 2016}, pages 2979--2985.

\bibitem[{Zhang and Qian(2020)}]{emnlp20-asc}
Mi~Zhang and Tieyun Qian. 2020.
\newblock \href {https://doi.org/10.18653/v1/2020.emnlp-main.286} {Convolution
  over hierarchical syntactic and lexical graphs for aspect level sentiment
  analysis}.
\newblock In \emph{Proceedings of the 2020 Conference on Empirical Methods in
  Natural Language Processing, {EMNLP} 2020}, pages 3540--3549.

\bibitem[{Zhang et~al.(2021)Zhang, Li, Deng, Bing, and Lam}]{acl21-gabsa}
Wenxuan Zhang, Xin Li, Yang Deng, Lidong Bing, and Wai Lam. 2021.
\newblock \href {https://doi.org/10.18653/v1/2021.acl-short.64} {Towards
  generative aspect-based sentiment analysis}.
\newblock In \emph{Proceedings of the 59th Annual Meeting of the Association
  for Computational Linguistics and the 11th International Joint Conference on
  Natural Language Processing, {ACL/IJCNLP} 2021}, pages 504--510.

\bibitem[{Zhao et~al.(2020)Zhao, Huang, Zhang, Lu, and Xue}]{acl20-spanmlt}
He~Zhao, Longtao Huang, Rong Zhang, Quan Lu, and Hui Xue. 2020.
\newblock \href {https://doi.org/10.18653/v1/2020.acl-main.296} {Spanmlt: {A}
  span-based multi-task learning framework for pair-wise aspect and opinion
  terms extraction}.
\newblock In \emph{Proceedings of the 58th Annual Meeting of the Association
  for Computational Linguistics, {ACL} 2020}, pages 3239--3248.

\bibitem[{Zhou et~al.(2015)Zhou, Wan, and Xiao}]{aaai15-acp}
Xinjie Zhou, Xiaojun Wan, and Jianguo Xiao. 2015.
\newblock \href {http://www.aaai.org/ocs/index.php/AAAI/AAAI15/paper/view/9764}
  {Representation learning for aspect category detection in online reviews}.
\newblock In \emph{Proceedings of the Twenty-Ninth {AAAI} Conference on
  Artificial Intelligence}, pages 417--424.

\end{thebibliography}
\bibliographystyle{acl_natbib}

\end{document}